\documentclass[10pt]{article}
\usepackage[margin=0.75in]{geometry}

\usepackage[pdftex]{graphicx}
\graphicspath{{../pdf/}{../jpeg/}}
\DeclareGraphicsExtensions{.pdf,.jpeg,.png}
\usepackage{amssymb,amsmath,url}
\usepackage{bm} 
\usepackage{multirow}
\usepackage{booktabs}	
\usepackage{color}
\usepackage{subcaption}
\usepackage{bigfoot}		
\usepackage{algpseudocode}
\usepackage{pifont}
\usepackage{tikz}
\usepackage{color}
\usepackage{cite}
\usepackage{setspace}
\usepackage{lipsum}
\usepackage{algorithm} 
\usepackage{algpseudocode}
\usepackage[pdftex]{graphicx}
\usepackage{epsfig}
\usepackage{comment}
\usepackage{enumitem} 
\usepackage{amssymb}
\usepackage{hyperref}
\usepackage{fontawesome5}

\ifodd 0
    \newcommand\revJnl[1]{{\color{blue}#1}}
    \newcommand{\comJnl}[1]{\textbf{\color{red} (COMMENT: #1)}} 
\else
    \newcommand\revJnl[1]{{#1}}
    \newcommand{\comJnl}[1]{}
\fi

\ifodd 0
    \newcommand\revLcss[1]{{\color{blue}#1}}
    \newcommand{\comLcss}[1]{\textbf{\color{red} (COMMENT: #1)}} 
\else
    \newcommand\revLcss[1]{{#1}}
    \newcommand{\comLcss}[1]{}
\fi

\algrenewcommand\algorithmicrequire{\textbf{Input:}}
\algrenewcommand\algorithmicensure{\textbf{Output:}}
\algrenewcommand\algorithmicforall{\textbf{For}}

\usepackage{amsthm} 
  
\newtheorem{proposition}{Proposition}

\allowdisplaybreaks

\newtheorem{remark}{Remark}

\newtheorem{assumption}{Assumption}
\newtheorem{example}{Example}

\newcommand{\E}{\mathbb{E}}

\DeclareMathOperator*{\argmax}{arg\,max}

\newcommand{\be}{\begin{equation}}
\newcommand{\ee}{\end{equation}}
\newcommand{\bee}{\begin{eqnarray}}
\newcommand{\eee}{\end{eqnarray}}

\allowdisplaybreaks

\newcommand{\bse}{\begin{subequations}}
	\newcommand{\ese}{\end{subequations}}

\renewcommand{\Pr}{\mathop{\mathbb{P}}}

\newcommand{\defeq}{\overset{\text{\tiny def}}{=}}

\ifodd 0
    \newcommand\rev[1]{{\color{magenta}#1}}
    \newcommand{\com}[1]{\textbf{\color{red} (COMMENT: #1)}} 
\else
    \newcommand\rev[1]{{#1}}
    \newcommand{\com}[1]{}
\fi

\ifodd 0 

\else
    
    \newcommand{\accom}[1]{}
\fi

\ifodd 0 
\newcommand{\cnjCom}[1]{{\textbf{\color{red} (CNJ$\to$ #1)}}}
\else
    
    \newcommand{\cnjCom}[1]{}
\fi


\title{\LARGE \bf 
Violation-Aware Contextual Bayesian Optimization for Controller Performance Optimization with Unmodeled Constraints
}
\author{Wenjie Xu\thanks{{Laboratoire d’Automatique}, École polytechnique fédérale de Lausanne, Lausanne, Switzerland.~{\faIcon{envelope}~\texttt{\{wenjie.xu, colin.jones\}@epfl.ch}}.} \footnotemark[2],   
Colin N Jones\footnotemark[1],  Bratislav Svetozarevic\thanks{Swiss Federal Laboratories for Materials Science and Technology, Switzerland.~\faIcon{envelope}~\texttt{bratislav.svetozarevic@empa.ch}},\\  Christopher R. Laughman\thanks{Mitsubishi Electric Research Laboratories, Cambridge, MA, USA.~\faIcon{envelope}~\texttt{laughman@merl.com}}, 
Ankush Chakrabarty\footnotemark[3] \footnotemark[4]\thanks{Corresponding author. \faIcon{envelope}~\texttt{achakrabarty@ieee.org}. \faIcon{phone-square-alt}~+1 (617) 758-6175. \faIcon{map-marker-alt} 201 Broadway, 8th Floor, Cambridge, MA 02139, USA.}
}
\allowdisplaybreaks
\begin{document}
\maketitle 

\begin{abstract}
We study the problem of  performance optimization of closed-loop control systems with unmodeled dynamics.  Bayesian optimization (BO) has been demonstrated to be effective for improving closed-loop performance by automatically tuning controller gains or reference setpoints in a model-free manner. However, BO methods have rarely been tested on dynamical systems with unmodeled constraints \revJnl{and time-varying ambient conditions}. In this paper, we propose a violation-aware \revJnl{contextual} BO algorithm (\revJnl{VACBO}) that optimizes closed-loop performance while simultaneously learning constraint-feasible solutions \revJnl{under time-varying ambient conditions}. Unlike classical constrained BO methods which allow unlimited constraint violations, or `safe' BO algorithms that are conservative and try to operate with near-zero violations, we allow budgeted constraint violations to improve constraint learning and accelerate optimization. We demonstrate the effectiveness of our proposed \revJnl{VACBO} method for energy minimization of industrial vapor compression systems \revJnl{under time-varying ambient temperature and humidity}.

\end{abstract}

\section{INTRODUCTION}
Closed-loop systems can often be optimized after deployment by altering controller gains or reference inputs guided by the performance observed through operational data. Manually tuning these control parameters often requires care and effort along with considerable task-specific expertise. Algorithms that can automatically adjust these control parameters to achieve optimal performance are therefore invaluable for saving manual effort, time, and expenditure. 

\revLcss{The optimal performance of a control system is generally defined via domain-specific performance functions whose arguments are outputs measured from the closed-loop system.} While the map from measurements to performance may be clearly defined, the map from control parameters (that can actually be tuned) to performance is often unmodeled or unknown, since closed-form system dynamics may not be available during tuning~\cite{chakrabarty2021_VCS}. It is thus common to treat the control parameters-to-performance map as a black-box, and \revLcss{design a data-driven tuning algorithm, where data is collected by experiments or simulations.} However, since both experimentation and high-fidelity software simulations are expensive, tuning algorithms must be designed to assign a near-optimal set of control parameters with as few experiments/simulations (equivalently, performance function evaluations) as possible. \revLcss{Therefore, existing data-driven methods that need a large number of samples, such as genetic algorithms~\cite{da2000application}, can be impractical.}

It is precisely for this reason that Bayesian optimization~(BO)\footnote{Also known as efficient global optimization~(e.g., in~\cite{jones1998efficient}) or kriging~(e.g., in~\cite{jeong2005efficient}) in optimization and engineering literature. } has received widespread attention in the context of closed-loop performance optimization. \revLcss{BO is a sample-efficient derivative-free global optimization method~\cite{jones1998efficient, xu2022lower} that utilizes probabilistic machine learning to intelligently search through parameter spaces. \cite{frazier2018tutorial} gives a detailed survey of Bayesian Optimization. 
In recent work, BO has demonstrated potential in controller gain tuning. 
For example, BO has been applied to the tuning of the PI controller of a heat pump~\cite{khosravi2019controller} and the tuning of PID cascade controller gains~\cite{konig2020safety}.
BO has also been applied to the performance optimization of model predictive control. For example, BO was applied to optimize the nominal linear model of a predictive controller~\cite{bansal2017goal}, to tune the parameters of MPC to optimize the closed-loop performance~\cite{piga2019performance}, and to generate candidate parameters for data-driven scenario optimization~\cite{paulson2020data}. BO has also been proposed to select closed loop kernel based model~\cite{beckers2019closed}. BO was used in many other various real-world control applications, such as wind energy systems parameter tuning~\cite{baheri2017altitude, baheri2020waypoint}, engine calibration~\cite{pal2020multi}, and space cooling system optimization~\cite{chakrabarty2021_VCS}.
}
\begin{figure*}[t]
     \centering
     \begin{subfigure}[b]{0.32\textwidth}
         \centering
         \includegraphics[width=\textwidth]{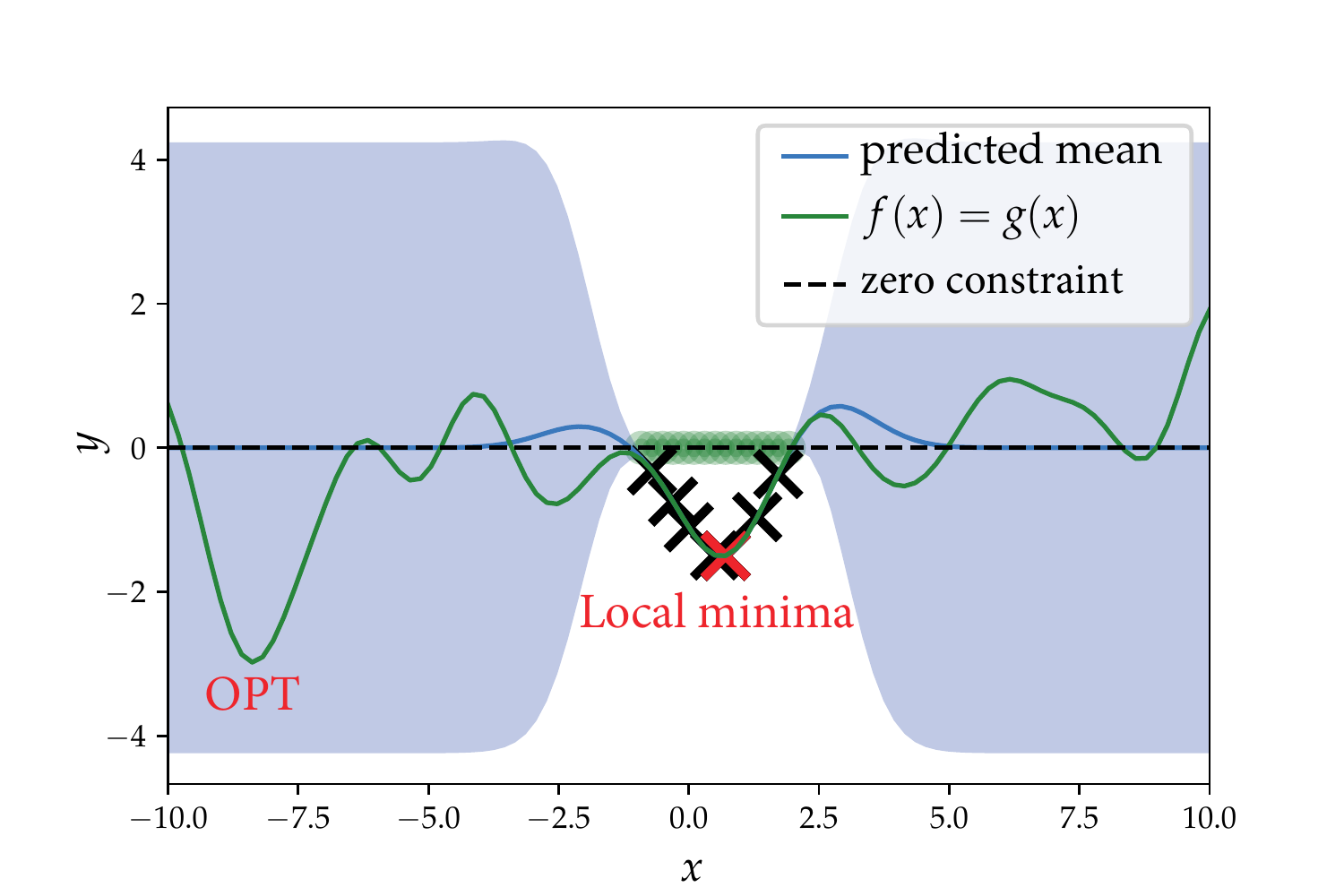}
         \caption{Safe BO~\cite{sui2015safe}}
         \label{fig:safe_bo_demo}
     \end{subfigure}
     \hfill
     \begin{subfigure}[b]{0.32\textwidth}
         \centering
         \includegraphics[width=\textwidth]{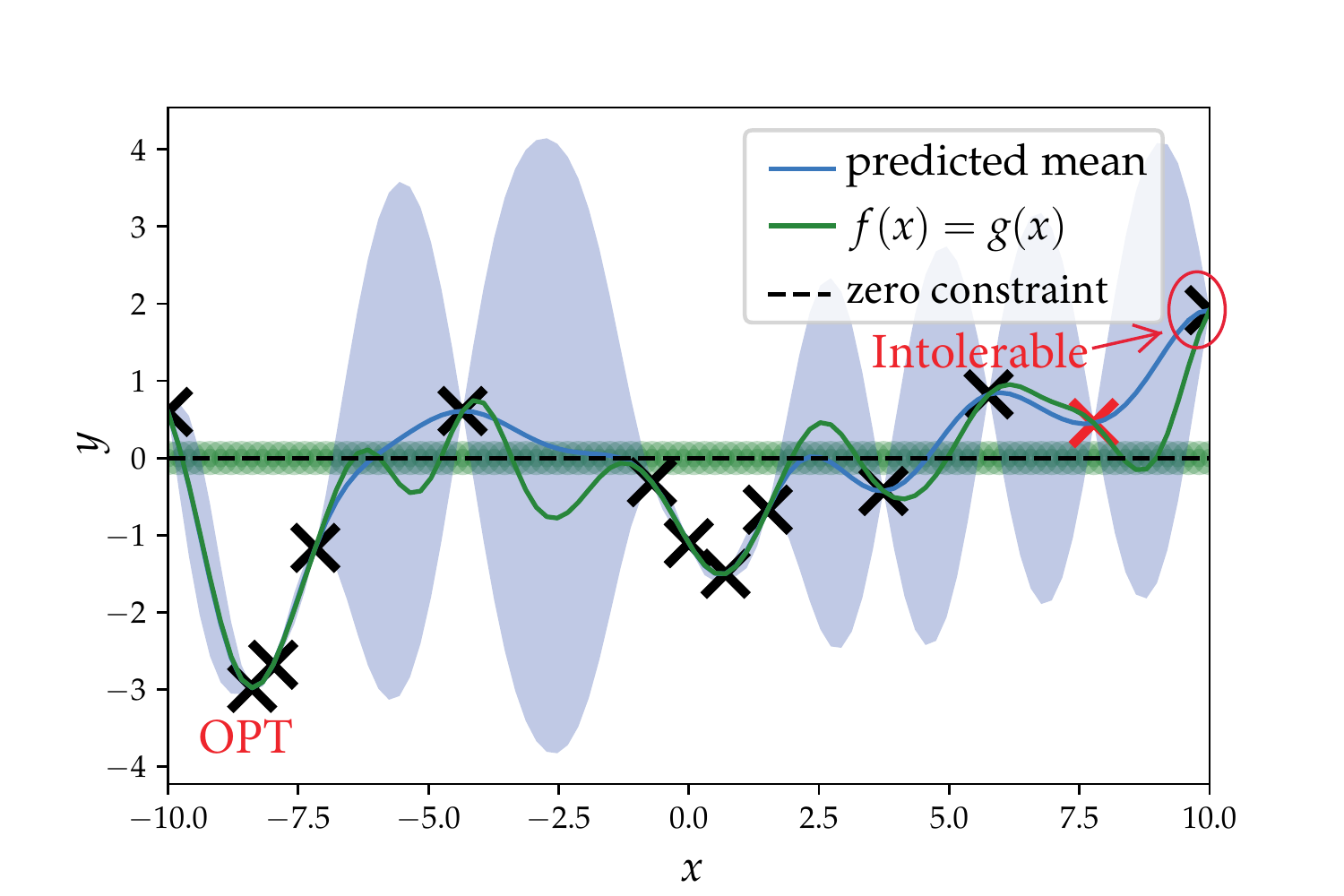}
         \caption{Constrained BO~\cite{gelbart2014bayesian}}
         \label{fig:c_bo_demo}
     \end{subfigure}
     \hfill
     \begin{subfigure}[b]{0.32\textwidth}
         \centering
         \includegraphics[width=\textwidth]{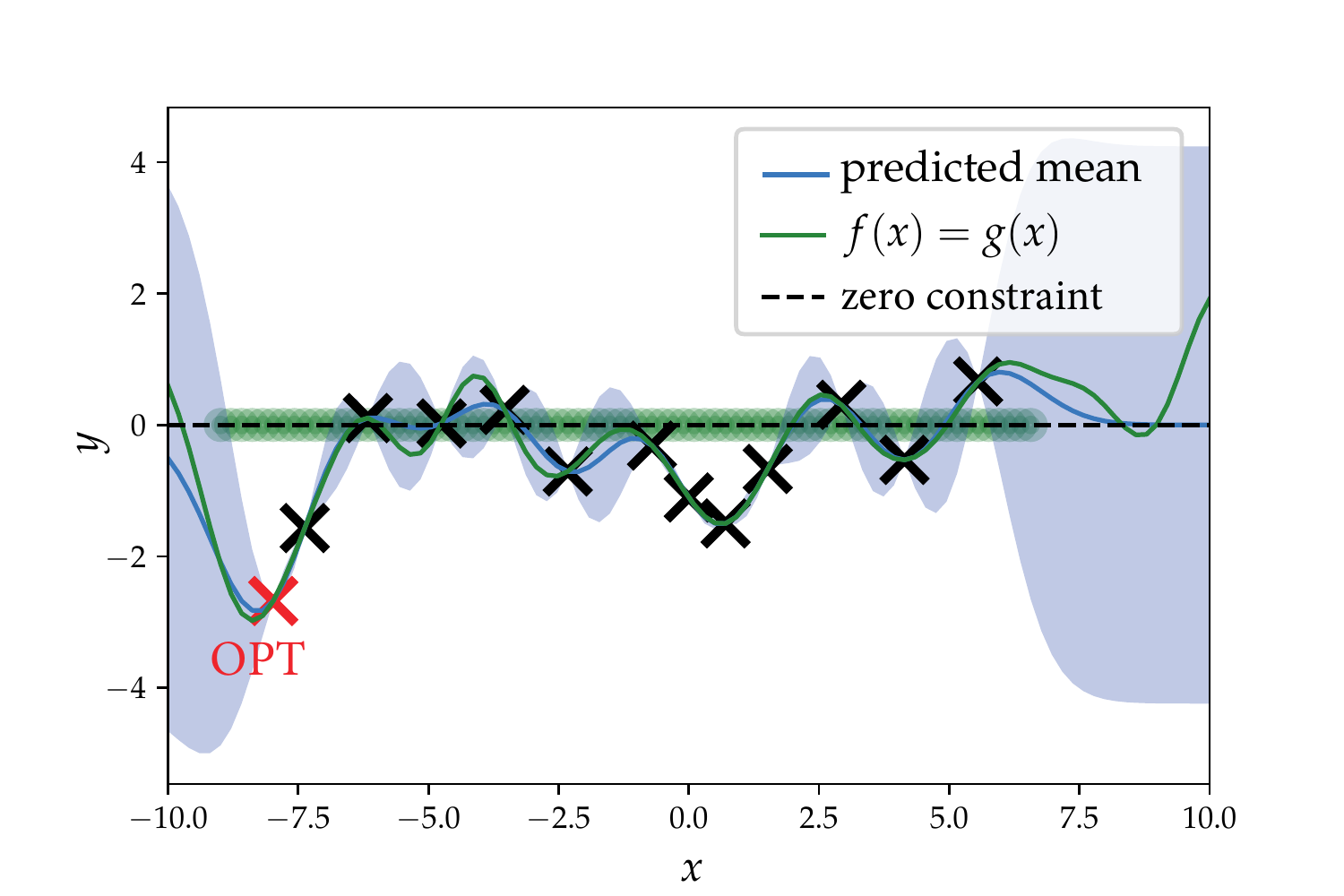}
         \caption{Our violation-aware BO}
         \label{fig:vabo_demo}
     \end{subfigure}
        \caption{\revJnl{A motivating example comparing our violation-aware BO to existing state-of-the-art methods. Safe BO gets stuck in a local minimum and fails to identify the global optimum, while generic constrained BO identifies the global optimum but can incur large constraint violations during the sampling process. In contrast, our method can simultaneously identify the global minimum and manages the constraint violation well.}}
        \label{fig:compare_demos}
\end{figure*}

A challenge that has garnered recent interest is that of \textit{safe} Bayesian optimization; that is, BO in the presence of safety-critical system constraints. These constraints may also be unmodeled (`black-box'), as a mathematical representation of the constraint with respect to the control parameters is not always known or straightforward to represent. \revLcss{To handle these constraints, \rev{safe BO methods have} been recently proposed \revLcss{in}~\revLcss{\cite{sui2015safe}}\revLcss{, improved in~\cite{sui2018stagewise}, and \revLcss{extended to a more general setting in~\cite{turchetta2020safe}}}. \revLcss{T}hese methods either operate on the principle of \revLcss{not allowing any} constraint violations during optimization, or leverage partial model knowledge to ensure safety via Lyapunov arguments~\cite{chakrabarty2021safe}.} In either case, safe BO \revLcss{learns} feasible optima without violating unmodeled constraints, or \rev{risks} their violation with a predefined small probability. Often, this conservativeness results in obtaining local minima, slow convergence speeds, and reduced data efficiency.
Conversely, generic constrained optimization with BO learns constraints without paying heed to the amount of constraint violation during the exploration phase~\cite{gardner2014bayesian, gelbart2014bayesian}. More recently, a group of works~\cite{xu2022constrained, xu2022config} propose an optimistic constrained optimization approach, with applications to control system tuning. These methods are mostly agnostic to the deleterious consequences of constraint violation, such as long-term damage to expensive hardware caused by large violations, rendering them impractical for many industrial applications. \revLcss{Another direction of BO research proposes the use of budgets on the cost of samples~(e.g., neural network training time~\cite{snoek2012practical}, wall clock time~\cite{lee2020cost}, sample number~\cite{lam2016bayesian} and system failures~\cite{marco2020excursion}). However, in these existing BO settings, the budget considered is usually related to the effort or failure risk for performance function evaluation, and does not provide a way to manage the magnitude of constraint violations.} 

For many industrial systems,  small constraint violations over a short period are often acceptable if that exploration \revLcss{improves the convergence rate of optimization}, but large violations are strongly discouraged. 
\revLcss{For example, in vapor compression systems (VCSs), it is imperative that constraint violation on variables such as compressor discharge temperature are limited to short time periods. We aim to find a set of near-optimal parameters within as few samples as possible since performance evaluation is time-consuming and available tuning time can be limited. Therefore, it may be desirable to systematically trade tolerable constraint violations for faster convergence and potentially skipping local minima. In other words, for VCSs, the benefits of accelerated global convergence  outweigh the cost of short-term constraint violations.} 

The performance of control systems are also influenced by exogenous signals such as variations in the environment. We refer to such signals as \textit{context} variables. It is often necessary to adapt control policies to maintain the performance and feasibility of the control system despite changes in the context variables. 
For instance, a controller designed for both performance and safety may adapt to prioritize safety over performance if the context variable predicts an unsafe event about to occur.
In order to systematically incorporate contextual information, contextual Bayesian optimization approaches have been proposed in~\cite{krause2011contextual}, where the inputs to the learner include the context variables augmented with the optimization variables.  Contextual Bayesian optimization approaches have recently been tested on controller design applications: for example, safe Bayesian optimization has been extended to the contextual case in~\cite{fiducioso2019safe} to tune a room temperature  controller via PID gain tuning. Furthermore, in~\cite{park2020contextual}, contextual Bayesian optimization is applied to cooperative wind farm control to maximize the efficiency of generated power. However, the incorporation of context in our setting, that is: how to simultaneously tune the controlled system efficiently and manage the constraint violations under a tolerable level with time-varying contexts, has not been studied previously, to the best of our knowledge.

In this paper, we propose a novel violation-aware contextual Bayesian optimization approach~(VACBO~\footnote{Code is available at \url{https://github.com/PREDICT-EPFL/VACBO}.}) that exhibits accelerated convergence compared to safe BO, while ensuring the violation cost is within a prescribed budget under time-varying contextual variables. We demonstrate that our VACBO algorithm is less conservative than `safe BO' algorithms that tend to be sample-inefficient and can get stuck in a local minimum because they cannot allow any constraint violation. The VACBO algorithm is also more cautious than constrained Bayesian optimization, which is agnostic to constraint violations and thus, is likely to incur large violation costs. Our VACBO algorithm is based on the principle of encouraging performance function evaluation at combinations of control parameters that greatly assist the optimization process, as long as it does not incur high constraint violations likely to result in system failure or irreversible damage. 

Our VACBO algorithm is an extension of the VABO algorithm~\cite{xu2021vabo} to the contextual case. We augment the input space with contextual variables and design a tractable auxiliary acquisition optimization problem specific to the contextual setting. \revJnl{More specifically, the VABO algorithm can not directly incorporate the impact of contextual variables, which can be significant in many applications. For example, the impact of ambient temperature and ambient humidity can be significant in vapor compression system set-point tuning. To incorporate the contextual variables, we augment the input variables, which we can control, with contextual variables that are measured from the environment. With this augmentation, our method can learn the joint impact of the input variables and contextual variables based on Gaussian process learning. Furthermore, the existing constrained expected improvement acquisition function used in VABO~\cite{xu2021vabo} and other constrained BO papers~\cite{gelbart2014bayesian, gardner2014bayesian} are not readily applicable to the contextual case. To address this issue, we extend the commonly used \textbf{C}onstrained \textbf{E}xpected \textbf{I}mprovement~(CEI) acquisition function to the contextual case and propose the \textbf{C}onstrained \textbf{P}roxy \textbf{E}xpected \textbf{I}mprovement~(CPEI) acquisition function.}

Our \textbf{contributions} include:

\begin{enumerate}
\item We propose a new variant of constrained BO methods for control parameter tuning that improves global convergence rates within a prescribed amount of constraint violation with guaranteed high probability under a time-varying contextual setting;

\item We propose a simple and tractable constrained auxiliary acquisition function optimization problem for trading off performance improvement and constraint violation; 

\item To incorporate the environmental conditions that impact the objective and constraints, we augment the input space by contextual variables and propose a new acquisition function by extending the commonly used \textsf{C}onstrained \textsf{E}xpected \textsf{I}mprovement~(\textsf{CEI})~\cite{gelbart2014bayesian,gardner2014bayesian} to the contextual setting; and,

\item We validate our algorithms on a set-point optimization problem using a high-fidelity VCS that has been calibrated on an industrial HVAC system, with ambient temperature and ambient humidity as two context variables. Simulation results with real-world weather signals as context variables demonstrate that our method efficiently minimizes the power consumption while simultaneously managing constraint violations within a tolerable level.
\end{enumerate}

\revJnl{We now state the organization of our paper. In Sec.~\ref{sec:prelim}, we present the statement of our problem and our proposed solution concept. Then in Sec.~\ref{sec:vacbo_alg}, we present our VACBO algorithm. After that, we give the application result of a case study on vapor compression system in Sec.~\ref{sec:case_study}. Finally, we conclude the paper in Sec.~\ref{sec:conclusion}. 

}

\section{Preliminaries}
\label{sec:prelim}
\subsection{Problem Statement}
We consider closed-loop systems of the form
\be
\label{eq:cl-sys}
\xi_+ = F(\xi, \theta, \revJnl{z}),
\ee
where $\xi,\xi_+\in\mathbb R^{n_\xi}$ denote the system state and \revJnl{the successor state} (respectively), $\theta\in\Theta\subset \mathbb R^{n_\theta}$ the control parameters (e.g., set-points) to be tuned, \revJnl{$z\in\mathcal Z \subset\mathbb{R}^{n_z}$ the context variables~(e.g., ambient temperature) that affect the dynamics}, and \revJnl{$F(\cdot,\cdot, \cdot)$} the closed-loop dynamics with initial condition $\xi_0$. We assume that the closed-loop system~\eqref{eq:cl-sys} is designed such that it is exponentially stable to a control parameter and context dependent equilibrium state $\xi^\infty(\theta, z)$ for every $\theta\in\Theta$. We further assume that $\xi^\infty(\cdot)$ is a continuous map on $\Theta$. 

To determine the system performance, we define a continuous cost function \revJnl{$\ell(\theta, z):\mathbb{R}^{n_\theta+n_z}\to\mathbb{R}$} to be minimized, which is an unknown/unmodeled function of the parameters \revJnl{$\theta$ and the context variables $z$}. This is not unusual: while $\ell$ may be well-defined in terms of system outputs, it is often the case that the map from control parameters \revJnl{and context variables} to cost remains unmodeled; in fact, $\ell$ may not even admit a closed-form representation on $\Theta$ \revJnl{and $\mathcal{Z}$}, c.f.~\cite{chakrabarty2021_VCS,burns2018proportional}.

We also define $N$ unmodeled constraints on the system outputs that require caution during tuning. \rev{The $i$-th such constraint is given by $g_i(\theta, z):\mathbb{R}^{n_\theta+n_z}\to\mathbb{R},i\in[N]$}, where the notation $[N]\defeq\{i\in\mathbb{N}, 1\leq i\leq N\}$; we assume each $g_i(\cdot, \cdot)$ is continuous on $\Theta\times\mathcal{Z}$. We assume that the cost function $\ell(\theta, z)$ and every constraint $g_i(\theta, z), \; i\in[N]$ can be ascertained, either by measurement or estimation, during the hardware/simulation experiment. \revJnl{We introduce the brief notation $g(\cdot, \cdot)\le 0\triangleq g_i(\cdot, \cdot)\le 0,\; i\in [N]$} and assume that an initial feasible set of solutions is available at design time.
\begin{assumption}\label{asmp:initial_set}
The designer has access to a non-empty safe set \revJnl{$S_0\subset\Theta\times\mathcal{Z}$} such that for any \revJnl{$(\theta, z)\in S_0$}, all constraints are satisfied; that is, \revJnl{$g(\theta, z)\le 0$ for every $(\theta, z)\in S_0$}.
\end{assumption}
While such an initial set \revJnl{$S_0$} can be derived using domain expertise, it is likely that \revJnl{$S_0$} contains only a few feasible \revJnl{$(\theta, z)$}, and at worst could even be a singleton set. \revJnl{Assumption~\ref{asmp:initial_set} is a common assumption in the literature of safe optimization (e.g.,~\cite{sui2015safe}). Without this assumption, it is possible that in the initial steps, we may already sample some points with large violations, which makes the problem unlikely to be tractable at all. Furthermore, we notice that in many applications~(e.g., vapor compression system set-points tuning), an initial set of feasible (maybe suboptimal) set-points are known based on domain knowledge. Therefore, Assumption~\ref{asmp:initial_set} is a necessary but not restrictive assumption for our problem.}

We cast the control parameter tuning problem as a black-box constrained optimization problem, formally described by 
\revJnl{
\bse\label{eqn:formulation}
\begin{align}
\min_{\theta\in\Theta}  \qquad& \ell(\theta, z),\label{eqn:obj} \\
\text{subject to:} \qquad & g_i(\theta, z)\leq 0,\quad \forall i\in[N].\label{eqn:constraint}
\end{align}
\ese
}
\revJnl{The contextual variable $z$ represents \revJnl{some quantities reflecting} the environmental conditions\revJnl{, which} we can measure but not directly control at each step. We give an example in the following.
\begin{example}
Ambient temperature and ambient humidity are two important contextual variables that impact the operation of the vapor compression system. Both of them can be measured but not directly controlled during the operations of the vapor compression systems.
\end{example}
\rev{Furthermore,} $z$ can be time-varying with its own dynamics. We use $z_t$ to denote the value of $z$ at time step $t$.}
Our \textbf{objective} is to solve the constrained optimization problem~\eqref{eqn:formulation} with limited constraint violations \rev{during the optimization process}. 
Since the constraints are assumed to be unmodeled and a \rev{limited set} of feasible solutions is known at design time, we \rev{do not expect a guarantee of zero constraint violation. The tolerable amount and duration of constraint violations are problem-dependent.} \com{Bratislav \& Wenjie: Not actually, it depends on the algorithm; e.g. SafeBO will not violate; If you sample conservatively, the sampling efficiency is bad. If we strategically sample, as in VABO, we may have some limited, but controlled constraint violation. This limit we formally introduce as a \textit{budget for allowed constraint violations}.  Why not exploit constraint violation to obtain better solutions / exploring better solutions. Try to address the question of encountering the associated risks of allowing the algorithm to violate the constraints slightly -- within the budget. But this needs to be allowed by the domain expert.}\rev{In some applications, such as vapor compression systems, small constraint violations over a short-term are acceptable, while large constraint violations are strongly discouraged.} \rev{In such cases,} instead of being overly cautious and ending up with suboptimal solutions, we allow small constraint violations \textit{as long as the resulting knowledge gathered by evaluating an infeasible (in terms of constraint violation) $\theta$ accelerates the optimization process or helps avoid local minima}.
\cnjCom{This paragraph seems too strong. The formulation we take does not necessarily penalize large violation like this. We just assume that there is a user-provided function that measures how bad a violation is. It could well be the case that the engineer sets it up so that a large, but short, violation is considered better than a short, but long, violation (e.g., as is very common when tuning batch controller for process control, when off-spec means that product must be thrown away - so large and small violations are the same, only violation time matters)}\com{Wenjie: I think our problem formulation can take into account the user-specific tolerance-level to different patterns of constraint violations. I agree that our formulation does not necessarily requires penalizes large violations a lot. But I think in many application scenarios, it's true~(e.g., VCS). } \com{Bratislav: We can address those differences in large and short violations and small and long, but, not in this formulation. It would be possible, but we need to introduce time dependent violation, and this we shall do in the journal paper, and here say for future work.}

\begin{remark}
Our formulation~\eqref{eqn:formulation} can also optimize batch processes over finite-time horizons, say $T_h$. This would involve defining the objective and constraints over a batch trajectory with stage loss \revJnl{$\ell(\theta, z):=\tfrac{1}{T_h} \int_0^{T_h} l(\tau,\theta, z) \,\mathrm{d}\tau$}.
\end{remark}

\subsection{Proposed Solution}
We propose a modified Bayesian optimization framework to solve the problem~\eqref{eqn:formulation} that is violation-aware: the algorithm automatically updates the degree of risk-taking in the current iteration based on the severity of constraint violations in prior iterations. Concretely, for an infeasible  $\theta$, the constraint violation cost is given by  
\revJnl{
\bee
\bar c_i(\theta, z) \triangleq c_i\left([g_i(\theta, z)]^+\right)\label{equ:violation_cost}, \quad i\in [N]
\eee
}
where \revJnl{$[g_i(\cdot, \cdot)]^+ := \max\{g_i(\cdot, \cdot), 0\}$ and $c_i:\mathbb{R}_{\ge 0}\to\mathbb{R}_{\ge 0}$}. \rev{Note that $g_i$ corresponds to physically meaningful system outputs that we can measure, e.g., temperature.}
This violation cost function $c_i$ is user-defined as a means to explicitly weigh the severity of `small' versus `large' constraint violations. While the function $c_i$ is at the discretion of the designer, it needs to satisfy the following mild assumptions in order to achieve desirable theoretical properties; see \S3.
\begin{assumption}
\label{assump:cost_func}
The violation \revJnl{cost} function $c_i$ satisfies:
\begin{enumerate}[label=(A\arabic*)]
\item $c_i(0)=0$,\label{enu:zero_vio_zero_cost}
\item $c_i(s_1)\geq c_i(s_2)$, if $s_1>s_2\geq 0$,\label{enu:mono}
\item $c_i$ is left continuous on $\mathbb{R}_{\ge0}$.
\end{enumerate}
\end{assumption}

Assumption~\ref{assump:cost_func} captures some intuitive properties required of the violation \revJnl{cost} function. According to (A1), there is no cost associated with no violation. From (A2), we ensure that the violation cost is monotonically \rev{non-decreasing} with increased violations. Finally (A3) ensures that this monotonic increase is smooth and does not exhibit discontinuous jumps from the left. 

\cnjCom{It's not clear that we need this function $c$, or that Assumption 2 does anything / is valid.\\[5pt]
The function $g$ is entirely user-defined. This means that it could be set to be $g_i(\theta) = c_i^{-1} \circ \hat g_i(\theta)$, which means that we can "undo" the action of $c_i$ through the selection of $g$, and thereby have $\bar c_i$ be anything at all. i.e., the assumption 2 can be met, and yet the map from $\theta$ to $\bar c$ can be anything.\\[5pt]
Because there's enough degrees of freedom in the selection of $g$ - we could just replace $c$ by $[g]^+$.\\[5pt]
All the properties listed in the paragraph above are also satisfied by $c = [g]^+$}
\com{Wenjie:Mathematically maybe true. But I think here $g$ has special physical meaning in different applications, like the tempratures, etc.. And $c_i$ is input from user, who may have a clear physical understanding how much to penalize different amount of violations. If we fix $c$ to be $[g]^+$, then we move the complexity to the design of $g$, which may not be straightforward for a user of our algorithm.}

To adapt the degree of risk-taking based on prior data obtained, we define a violation budget over a horizon of $T\in\mathbb N$ optimization iterations.
\revJnl{Our goal is to} sequentially search over $T$ iterations $\{\theta_t\}_{t=1}^T$ while \rev{using a prescribed budget} of constraint violations in order to obtain a feasible and optimal set of parameters
\revJnl{\begin{align}\label{eq:budgeted_optimization_problem}
\notag  \min_{\substack{\theta_t\in\Theta;\\ g(\theta_t, z_t)\leq 0}} &\ell(\theta_t, z_t)\;\; \\
\textrm{subject to:}\;\;&
\sum_{t=1}^{T}\bar c_i(\theta_t, z_t)\leq B_i\enspace,\quad i\in[N]
\end{align}
}
where $B_i$ denotes a budget allowed for the $i$-th violation cost. 
Note that this formulation is a generalization of  well-known constrained\revLcss{/safe} Bayesian optimization formulations proposed in the literature. \revLcss{If we set all $B_i\equiv 0$, then our  formulation is \revJnl{closely} related to safe BO~\cite{sui2015safe,sui2018stagewise}. Alternatively, setting $B_i\equiv \infty$ reduces our problem to constrained BO agnostic to violation cost~\cite{gardner2014bayesian, gelbart2014bayesian}.}

\section{Violation-Aware Contextual Bayesian Optimization}
\label{sec:vacbo_alg}
\subsection{Bayesian Optimization Preliminaries}
For Bayesian optimization, one models $\ell(x)$ and $g(x)$ as functions sampled from independent Gaussian processes. In our case, the input $x$ to the Gaussian process consists of control parameters $\theta$ and the context variables $z$. At iteration $t$, conditioned on previous input and function evaluation data $\mathcal{D} := \left\{(\theta, z)_{1:t}, \ell((\theta, z)_{1:t})\right\}$, the posterior mean and standard deviation of $\ell$ is given by
$$
 \mu_\ell(x|\mathcal D) = k_\ell^\top(x, x_{\mathcal D}) K_\ell^{-1}\Delta y_\ell + \mu_{\ell,0}(x)$$ and
$$\sigma^2_\ell(x|\mathcal{D})=k_\ell(x, x)-k_\ell^\top(x, x_{\mathcal{D}})K_\ell^{-1}k_\ell(x_{\mathcal{D}}, x),$$
where $x_{\mathcal{D}}=(\theta, z)_{1:t}$ is the set of control parameters and context variables with which previous experiments/simulations have been performed. Here,
\begin{align*}
k_\ell(x, x_{\mathcal{D}}) &\triangleq[k_\ell(x, x_i)]_{x_i\in x_{\mathcal{D}}},\\
k_\ell(x_{\mathcal{D}}, x) &\triangleq[k_\ell(x_i, x)]_{x_i\in x_{\mathcal{D}}},\\
K_\ell &\triangleq \left(k_\ell(x_{i},x_{j})\right)_{x_i, x_j\in x_\mathcal{D}},\\
\Delta y_\ell &\triangleq[\ell(x_i)-\mu_{\ell,0}(x_i)]_{x_i\in x_\mathcal D},
\end{align*}
and $k_\ell(\cdot, \cdot)$ is a user-defined kernel function and $\mu_{\ell,0}$ is the prior mean function, both associated with $\ell$; see~\cite{frazier2018tutorial} for more details on kernel and prior mean selection. The above quantities are all column vectors, except $K_\ell$, which is a positive-definite matrix.
For the constraint functions $g$, similar expressions for the posterior mean $\mu_{g_i}(x|\mathcal{D})$ and standard deviation $\sigma_{g_i}(x|\mathcal{D})$ can be obtained.

The kernelized functions above provide tractable approximations of the cost of the closed-loop system, along with the constraint functions, both of which were hitherto unmodeled/unknown. Classical BO methods use the statistical information embedded within these approximations to intelligently explore the search space $\Theta$ via acquisition functions. A specific instance of an acquisition function commonly used in constrained BO is the constrained expected improved (CEI) function~\cite{gardner2014bayesian}. It is defined as the expectation of the multiplication of improvement as compared to the incumbent best objective sampled so far and the feasibility indicator. However, in the contextual case, the objective may heavily rely on context. The incumbent best objective value under a favorable context may mislead the parameter search under the adversarial context. 

Therefore, instead of using the incumbent best objective, we propose a two-step approach. In the first step, we use the minimum value of the posterior mean of $\ell$ to construct a proxy of the best objective sampled so far under a different context as in~\eqref{equ:prox_min}.   
\begin{align}
\hat{\ell}_t^{\min}(z) = &\min_{\theta\in\Theta} \mu_\ell((\theta, z)|\mathcal D)
\label{equ:prox_min}
\end{align}
In the second step, we use the minimum value of posterior mean $\hat{\ell}_t^{\min}$ to construct a new acquisition function, the \textsf{C}onstrained \textsf{P}roxy \textsf{E}xpected \textsf{I}mprovement~($\mathsf{CPEI}$), given by
\begin{equation}
\mathsf{CPEI}((\theta, z)|\mathcal{D}) = \E\left(\prod_{i\in[N]}\mathbf{1}_{g_i(\theta, z)\leq 0}\; I(\theta, z)|\mathcal D\right),
\label{equ:eic_def}
\end{equation}
where $\mathbf{1}$ denotes the indicator function, $\mathbb E$ denotes the expectation operator, and
$I(\theta, z)=\max\{0, \hat{\ell}^{\min}_t(z)-\ell(\theta, z)\}$
is the improvement of $(\theta, z)$ over the proxy $\hat{\ell}^{\min}_t(z)$ for the best incumbent solution over $t$ iterations. 

As $g_i(\theta, z),\forall i\in[N]$ and $\ell(\theta, z)$ are independent, we deduce
\begin{equation}\label{equ:cei_first}
\mathsf{CPEI}(\theta, z|\mathcal{D})=\prod_{i\in[N]} \Pr(g_i(\theta, z)\leq 0|\mathcal{D})\E\left(I(\theta, z)|\mathcal{D}\right).
\end{equation}
We have
$\Pr(g_i(\theta, z)\leq0|\mathcal{D})=\Phi\left(\tfrac{-\mu_{g_i}(\theta, z|\mathcal{D})}{\sigma_{g_i}(\theta, z|\mathcal{D})}\right)$,
and the closed-form expression of expected improvement~\cite{jones1998efficient},
\bee\label{eq:acqfn_cei_b}
\E\left(I(\theta, z)|\mathcal{D}\right)=\Delta l(\theta, z|\mathcal{D}) \Phi\left(w\right)+\sigma_\ell(\theta, z|\mathcal{D})\phi\left(w\right),
\eee
where $\Delta l(\theta, z|\mathcal{D})=\hat{\ell}^{\min}_t(z)-\mu_{\ell}(\theta, z|\mathcal{D})$,  $w=\tfrac{\hat{\ell}_t^{\min}(z)-\mu_\ell(\theta, z|\mathcal{D})}{\sigma_\ell(\theta, z|\mathcal{D})}$, $\Phi(\cdot)$ and $\phi(\cdot)$ are the standard normal cumulative distribution and probability density functions, respectively. 

\subsection{VACBO Algorithm}

Our VACBO algorithm proposes an auxiliary optimization problem that leverages the constrained proxy expected improvement acquisition function to guide the search of feasible points with potentially lower objective to evaluate while ensuring (with high probability) that the violation cost will remain within a prescribed budget.

Given the total violation cost budget, a question is how to allocate the budget across different samples. Intuitively, it may be beneficial to dynamically adjust the violation cost budget allocated to a single step. For example, if we find that we incur no violation cost for several steps, it is possible that we are overly cautious in those steps and may get stuck in a local minimum. So we can then increase the violation cost budget allocated for the next step. It is also possible that we incur significant violation in some step due to the over-confidence in the constraint function prediction by Gaussian process regression. In this case, we need to decrease the violation cost budget allocated to one single step. To capture this intuition in our algorithm, we design a violation cost budget allocation scheme to dynamically adjust the violation cost allocated to one single step. We use $B_{i,t}$ to denote the violation cost budget allocated to the step $t$ for the $i$-th constraint.
Our violation cost budget allocation scheme is given as,
\begin{equation}
\label{eq:budget_alloc_scheme}
B_{i,t} \triangleq \min\left\{\max\left\{B_i S_{i,t}-\sum_{\tau=1}^{t-1}\bar c_i(\theta_\tau, z_\tau), 0\right\}, B_i^{\max}\right\},
\end{equation}
where $S_{i,t}$ is a non-negative and non-decreasing sequence that satisfies $S_{i, T}=1$ and $B_i^{\max}$ is the maximum violation cost tolerable for the $i$-th constraint, which is a user-provided parameter that captures the user's maximum tolerance for constraint violation in one single step.

At this iteration, after observing the context variables, we solve the following \textit{auxiliary} problem 
\bse
\label{eqn:acquisition_problem}
\begin{align}
\theta^\star_t := \arg\max\limits_{\theta\in\Theta}\;&\mathsf{CPEI}(\theta, z|\mathcal{D}),\label{eqn:acq_obj} \\
\text{subject to:} & \prod_{i\in[N]}\Pr(\bar{c}_i(\theta, z)\leq B_{i,t})\geq 1-\epsilon_t, \label{eqn:acq_constraint}
\end{align}
\ese
to compute the next control parameter candidate $\theta_t^\star$, where $0<\epsilon_t\ll 1$
determines the probability of large constraint violation. Note that~\eqref{eqn:acq_obj} involves maximizing the constrained expected improvement type objective, which is common to cBO algorithms; c.f.~\cite{gardner2014bayesian}. Our modification using the budget, as written in~\eqref{eqn:acq_constraint}, enforces that the next sampled point will not use up more than 
the violation cost budget $B_{i,t}$ for all constraints with a probability of at least $1-\epsilon_t$, conditioned on the data seen so far. This modification allows us to trade a prescribed level of violation risk for more aggressive exploration, leading to faster convergence. Although we use the constrained proxy expected improvement acquisition function here, we can easily generalize to other acquisition functions by simply replacing $\mathsf{CPEI}$ in the objective~\eqref{eqn:acq_obj}.

We now discuss how to efficiently solve the auxiliary problem~\eqref{eqn:acquisition_problem}. Recall from Assumption~\ref{assump:cost_func} that $c_i$ is non-decreasing on $\mathbb{R}_{\ge 0}$ for every $i\in[N]$. Therefore, we can define an inverse violation function
$c_i^{-1}(s) = \sup\{r\in\mathbb{R}_{\ge 0} \mid c_i(r)\leq s\}$
for any $s\in\mathbb{R}_{\ge 0}$.
Therefore, we can write $\Pr(\bar c_i(\theta, z)\leq B_{i,t}|\mathcal{D})=\Pr([g_i(\theta, z)]^+\leq c_i^{-1}(B_{i,t})|\mathcal{D})=\Pr(g_i(\theta, z)\leq c_i^{-1}(B_{i,t})|\mathcal{D})$.
Since $g_i(\theta, z)$ follows a Gaussian distribution with mean $\mu_{g_i}(\theta, z|\mathcal{D})$ and variance $\sigma_{g_i}^2(\theta, z|\mathcal{D})$, we get
\[
\Pr(\bar c_i(\theta, z)\leq B_{i,t}|\mathcal{D})=\Phi\left(\frac{c_i^{-1}(B_{i,t})-\mu_{g_i}(\theta|\mathcal{D})}{\sigma_{g_i}(\theta|\mathcal{D})}\right).
\]
When the number of control parameters $n_\theta$ 
is small (e.g., $<6$), we can place a grid on $\Theta$ and evaluate the cost and constraints of~\eqref{eqn:acquisition_problem} at all the grid nodes. The maximum feasible solution can then be used as the solution to the auxiliary problem. When the number of control parameters is large~(e.g., $n_\theta>6$), we can use gradient-based methods with multiple starting points to solve problem~\eqref{eqn:acquisition_problem}, since evaluating the learned GPs approximating $\ell$ and $g$ require very little computational time or effort when $T$ is not large (empirically, $<2000$).
We provide pseudocode for implementation in Algorithm~\ref{alg:VACBO}.

The following proposition provides a probabilistic guarantee of 
keeping the violation cost below the given budget. It highlights the ``violation-awareness'' property exhibited by VACBO.
\begin{proposition}\label{thm:vio_aware}
Fix $\delta\in (0,1)$ and  $T\in\mathbb N$.
If $\epsilon_t, t\in [T]$ are chosen such that ${\delta} = 1-\prod_{t=1}^T (1-\epsilon_t)$, then the VACBO algorithm satisfies the probability that
\[
\left\{\max_{t\in[T]}\bar c_i(\theta_t, z_t)\leq B_i^{\max} \textrm{ and } \sum_{t=1}^{T}\bar c_i(\theta_t, z_t)\leq B_i,\forall i\in[N]\right\}
\]
is at least $1-\delta$.
\end{proposition}
\begin{proof}
{
Let
$$
\mathcal{E}_t^1:=\left\{\max_{\tau\in[t]}\bar c_i(\theta_\tau, z_\tau)\leq B_i^{\max},\forall i\in[N]\right\}
$$

$$
\mathcal{E}_t^2:=\left\{\sum_{\tau=1}^{t}\bar c_i(\theta_\tau, z_\tau)\leq B_iS^i_t,\;\forall i\in[N]\right\}
$$

$$\mathcal{E}_t:=\mathcal{E}_t^1\cap\mathcal{E}_t^2,$$
where $t\in[T]$. Furthermore, we let
$$
\Delta\mathcal{E}_t^1:=\left\{\bar c_i(\theta_t, z_t)\leq B_i^{\max},\forall i\in[N]\right\}
$$

Notice that $\mathcal{E}_{t+1}^1=\mathcal{E}_t^1\cap \Delta\mathcal{E}_{t+1}^1$.
We have 
\begin{align}
\Pr\left(\mathcal{E}_T\right)&\geq\Pr\left(\mathcal{E}_{T-1}\right)\Pr\left(\mathcal{E}_{T}\mid \mathcal{E}_{T-1}\right)\nonumber\\
&=\Pr\left(\mathcal{E}_{T-1}\right)\Pr\left(\mathcal{E}_{T-1}^1\cap\Delta\mathcal{E}_T^1\cap\mathcal{E}_T^2|\mathcal{E}_{T-1}\right)\nonumber\\
&=\Pr\left(\mathcal{E}_{T-1}\right)\Pr\left(\Delta\mathcal{E}_T^1\cap\mathcal{E}_T^2|\mathcal{E}_{T-1}\right)\nonumber\\
&= \Pr\left(\mathcal{E}_{T-1}\right)\Pr\left(\bar c_i(\theta_T, z_T)\leq B_{i,T},\forall i\in[N]|\mathcal{E}_{T-1}\right)\nonumber\\
&\geq \Pr(\mathcal{E}_{T-1})(1-\epsilon_T),
\end{align}
where the last equality follows by that $$B_{i,T} \triangleq \min\left\{B_i S_{i,T}-\sum_{\tau=1}^{T-1}\bar c_i(\theta_{\tau}, z_\tau), B_i^{\max}\right\}$$ conditioned on $\mathcal{E}_{T-1}$.
By recursion, we have
$$\Pr\left(\mathcal{E}_T\right)
\geq\Pr\left(\mathcal{E}_1\right)\prod_{t=2}^{T}(1-\epsilon_t)\geq\prod_{t=1}^{T}(1-\epsilon_t)\nonumber=1-\delta,$$which concludes the proof since $S_T^i=1$.
}
\end{proof}

\begin{algorithm}[!ht]
	\caption{Violation-Aware Contextual Bayesian Optimization}\label{alg:VACBO}
	\begin{algorithmic}[1]
	\normalsize
	\State \textbf{Require}: VACBO horizons $T$, violation total budget $B_i,\forall i\in[N]$, maximum allowed violation cost for a single step $B_i^{\max}, \forall i\in[N]$, and an initial safe set of points $\mathcal{X}_0$ 
	\State Evaluate $\ell(\theta, z)$, $g_i(\theta, z),\forall i\in[N]$ for $(\theta, z)\in \mathcal{X}_0$ by performing experiments or simulation, or using historical data
    \State Initialize dataset $$\mathcal D=\left\{(\theta_t, z_t), \ell(\theta_t, z_t), g(\theta_t, z_t)\; \forall(\theta_t, z_t)\in \mathcal{X}_0\right\}$$
    \vspace{-3ex}	
	\For{$t\in[T]$}
	    \State Observe the context variables $z_t$ for step $t$
	    \State $B_{i,t} \triangleq \min\left\{\max\left\{B_iS_{i,t}-\sum_{\tau=1}^{t-1}\bar c_i(\theta_{\tau}, z_\tau), 0\right\}, B_i^{\max}\right\}$
	    \State $\Theta_t=\left\{\prod_{i\in[N]}\Pr(\bar c_i(\theta, z_t)\leq B_{i,t}|\mathcal{D})\geq 1-\epsilon_t|\theta\in\Theta\right\}$
    \State $\theta_{t}^\star=\argmax_{\theta\in \Theta_t}\mathsf{CPEI}(\theta, z_t|\mathcal{D})$ \Comment{Solving~\eqref{eqn:acquisition_problem}}
	    \State $\ell(\theta_t^\star, z_t), g(\theta_t^\star, z_t)\leftarrow$  perform experiment with $\theta_t^\star$ under the context $z_t$ 
	    \State Update dataset, $$\mathcal D\leftarrow \mathcal D \cup \{(\theta_t^\star, z_t), \ell(\theta_t^\star, z_t), g(\theta_t^\star, z_t)\}$$
	    \vspace{-3ex}
	    \State {Update Gaussian process posterior} 
	\EndFor
	\end{algorithmic}
\end{algorithm}

\section{Case Study: \revLcss{Constrained VCS Optimization}}
\label{sec:case_study}

\begin{figure*}[!t]
    \centering
    \includegraphics[width=0.9\columnwidth]{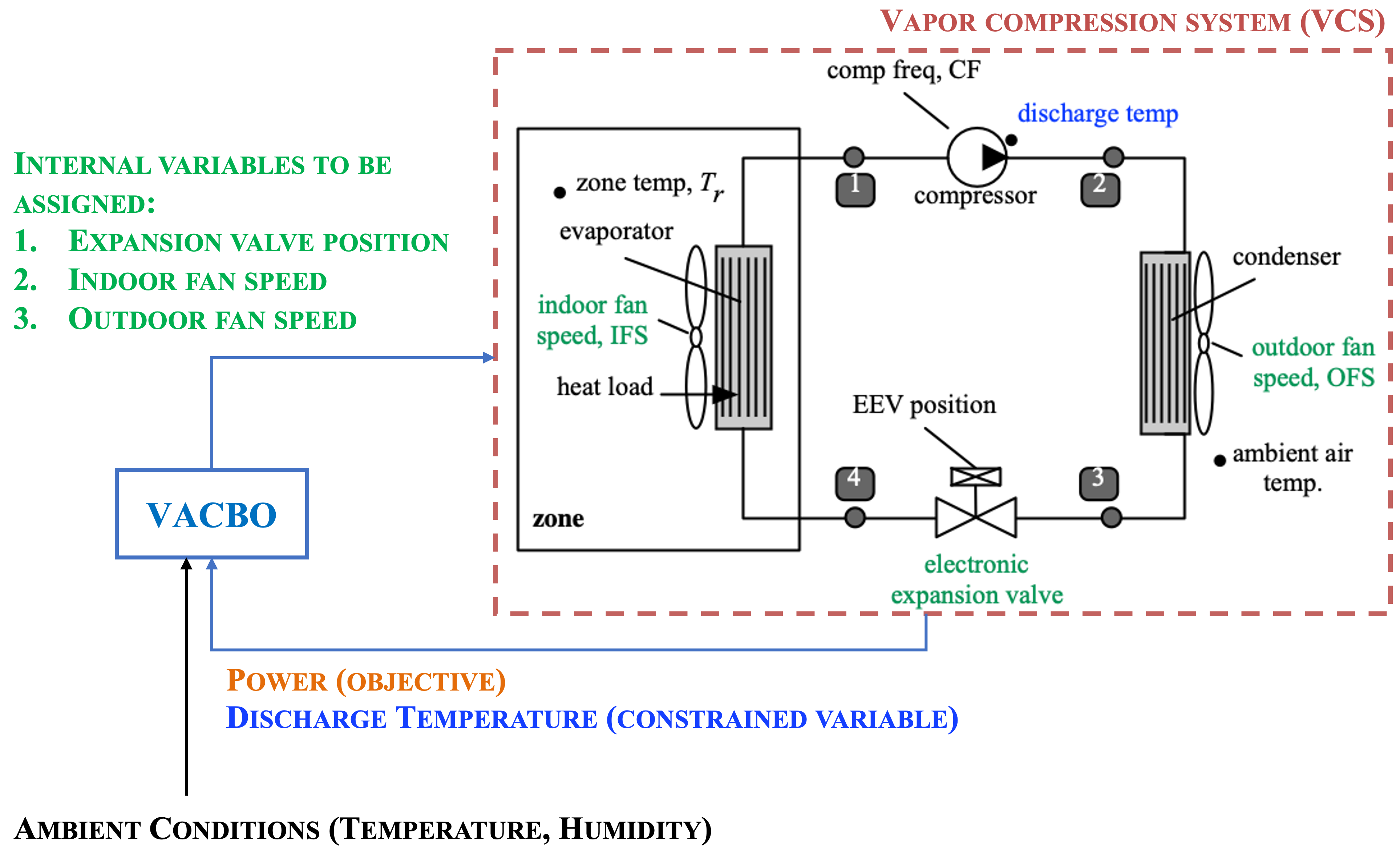}
    \caption{Schematic diagram of vapor compression system with our proposed VACBO controlling the EEV~(electronic expansion valve) and the two fans' speed. For simplicity, we do not show other measurements and controls.}
    \label{fig:vcs_vabo_diagram}
\end{figure*}

\revJnl{In this section, we apply the violation-aware contextual Bayesian optimization  framework to safely tune the set-points of a vapor compression system~(VCS)}.
As shown in Fig.~\ref{fig:vcs_vabo_diagram}, a VCS typically consists of a compressor, a condenser, an  expansion valve, and an evaporator. 
While physics-based models of these systems can be formulated as large sets of nonlinear differential algebraic equations to predict electrical power consumption, there are a variety of challenges in developing and calibrating these models. 
This motivates interest in directly using measurements of the power under different operating conditions to search for optimal set-points to the VCS actuators using data-driven, black-box optimization methods such as BO, to minimize the power consumption. 

\revJnl{
During the tuning process, one constraint, considered here, is on the temperature of the refrigerant leaving the compressor, also referred to as the `discharge temperature'~(see Fig.~\ref{fig:vcs_vabo_diagram}).} 
The discharge temperature must be managed because compressors are designed to operate within specific temperature ranges; excessively high temperatures can result in the breakdown of refrigerant oils and increase wear and tear.  In addition, high temperatures are often correlated with high pressures, which can cause metal fatigue and compromise the integrity of the pressurized refrigerant pipes in extreme cases. 
While managing the constraints mentioned above in the long run is critical, we also observe that small violations over short periods of time have limited harmful effects. Indeed, it may be beneficial to take the risk of short-period limited violation to accelerate the tuning process. 

\revJnl{Meanwhile, another challenge of tuning comes from the time-varying ambient conditions. Two major ambient conditions that have significant impact on the performance of vapor compression systems are the ambient temperature and the ambient humidity. The feasibility and optimality of VCS setpoints may be strongly correlated with these ambient conditions. For example, one set of fixed setpoints may lead to a feasible discharge temperature with low ambient temperature, but result in discharge temperature violation when the ambient temperature becomes high. It is necessary to adapt the set-points to the ambient conditions.}

\revJnl{
We cast the VCS optimization problem in the same form as ~\eqref{eqn:formulation}, with $\ell((\theta, z))$ denoting the steady-state power of the VCS with set points $\theta$ and contextual variables $z$. The constraint $g\le 0$ is given by $T_d((\theta, z))-\hat T_d\leq0$, where \rev{$T_d((\theta, z))$ is the steady-state discharge temperature with set points $\theta$ and context variables $z$, and} $\hat T_d$ is a safe upper bound. We close a feedback loop from compressor frequency to room temperature, leaving the set of 3 tunable set points $\theta$ as the expansion valve position and the indoor/outdoor fan speeds. We set the contextual variables to be ambient temperature and the ambient humidity, which are the two major ambient factors influencing the performance of vapor compression systems. The effects of these set points and contextual variables on power and discharge temperature are not easy to model, and no simple closed-form representation exists. In practice, we assign a setpoint $\theta$ under contexts $z$, wait for an adequate amount of time until the power signal resides within a 95\% settling zone, and use that power value as $\ell((\theta, z))$ and the corresponding discharge temperature as $T_d((\theta, z))$.
}


\subsubsection*{Implementation Details}
\revLcss{We use a high-fidelity model of the dynamics of a prototype VCS\footnote{Note that while the behavior of this model has been validated against \rev{a} real VCS, the numerical values and/or performance presented in this work are not representative of any product.} written in the \texttt{Modelica} language~\cite{modelica2017a} to collect data and optimize the set-points on-the-fly. A complete model description is available in~\cite{chakrabarty2021_VCS}.
The model was first developed in the \texttt{Dymola}~\cite{dymola2020a} environment, and then exported as a functional mockup unit (FMU)~\cite{fmi2019a}. Its current version comprises 12,114 differential equations.}
\revJnl{
We sample the system state each second. To leave enough time for the system state to converge to the steady state, we update the set-points every $180\textrm{s}$.}
Bayesian optimization is implemented in \texttt{GPy}~\cite{gpy2014}.

\revJnl{
We define our set-point search space \rev{$\Theta := [200, 350]\times[300, 450]\times[500, 850]$}, in expansion valve counts, indoor fan rpm~(\revJnl{revolutions per minute}), and outdoor fan rpm, respectively. We aim to keep the discharge temperature below $\hat T_d=333$~K; these constraints are set according to domain knowledge~\cite{burns2018proportional}. We initialize the \rev{simulator} at an expansion valve position of 340~counts, an indoor fan speed of 440~rpm, and an outdoor fan speed of 840~rpm, which is known to be a feasible set-point based on experience.
}

%
Constraint violations are penalized with the function $c_i(s)=s^2,\, s\in\mathbb{R}^+$. The quadratic nature of the violation cost implies that minor violations are not as heavily penalized as larger ones. The reason for this is that small violations over a small period of time are unlikely to prove deleterious to the long-term health of the VCS, whereas large violations could have more significant effects, even over short periods of time; for instance, damage to motor winding insulation or exceeding mechanical limits on the pressure vessel of the compressor. \revLcss{These constraints have been incorporated into the selection of the thresholds $\hat T_d$. Of course, the threshold values and a parameterized violation cost could be considered to be hyperparameters, and could be optimized via further experimentation.}
We choose the RBF kernel for our problem, which is commonly used in Bayesian optimization~\cite{frazier2018tutorial}, and compare our method to \revJnl{two other state-of-the-art BO methods, namely,} safe BO~\cite{berkenkamp2016safe, fiducioso2019safe} and generic \revJnl{constrained BO}~(cBO)~\cite{gelbart2014bayesian}. \revJnl{To ensure a fair comparison, we also extend the other two state-of-the-art BO methods to the contextual setting. More specifically, we augment the input space of Gaussian processes with contextual variables and inherit the algorithms of both safe BO and generic cBO.} \revLcss{Based on our domain experience \revJnl{and prior knowledge}, we set the kernel variance to be $15.0$~($2.0$, respectively), the kernel lengthscales \revJnl{in control parameters} to be $[50, 60, 70]$~($[20, 24, 28]$, respectively) for the objective~(constraint) and the kernel lengthscales \revJnl{in contextual variables} to be $[1.0, 0.06]$~($[1.0, 0.06]$, respectively) for the objective~(constraint).} 

\revJnl{We showcase the effect of varying $B_i, i=1$ by selecting $B_1=0$, $10$, and $20$, and set $S_{1,t}=a_{i} + b_i\frac{t}{T}$, where $a_i+b_i=1$ and $a_i\geq0$ and $b_i\geq0$.}
This choice allows the algorithm to use an increasing fraction of the \revJnl{total budget minus the violation cost incurred in the previous steps} when approaching the sampling limit $T$. \revJnl{Intuitively, such a choice of budget sequence gradually increases the violation risk level without sampling too aggressively in the initial several steps and allows us to reduce the budget based on the violation cost in the previous steps}. 
\revLcss{Proposition~\ref{thm:vio_aware} gives a conservative way of choosing $\epsilon_t$. In this case study, setting $\epsilon_t$ to a small constant $0.01$ works well.} \revJnl{We use ``VACBO $B$'' to indicate violation aware contextual BO with budget $B_1=B$.
}

\revJnl{
\emph{Generation of Context Sequences.} We consider both artificial recurring contexts and real-world contexts. In Sec.~\ref{sec:art_rec_cts}, we use artificially generated recurring contexts to showcase the effectiveness of our algorithm in optimizing the power consumption while managing the discharge temperature violation well. In Sec.~\ref{sec:real_cts}, we apply our algorithm to more realistic setting with real-world historic context sequences measured in Zurich, Switzerland.          
}

\subsection{Artificial Recurring Contexts}
\begin{figure}[htbp]
    \centering
    \includegraphics{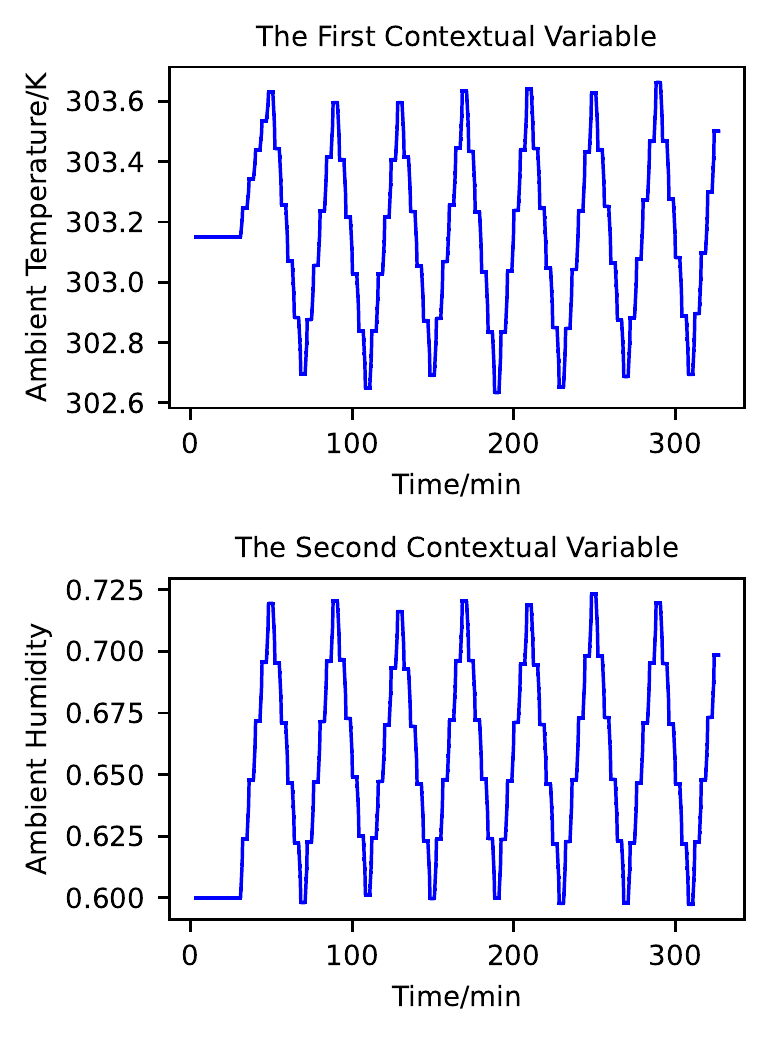}
    \caption{Recurring contexts used in our experiment.}
    \label{fig:recur_context}
\end{figure}

\begin{figure}[htbp]
\centering
\begin{subfigure}[h]{0.5\columnwidth}
    \centering
    \includegraphics{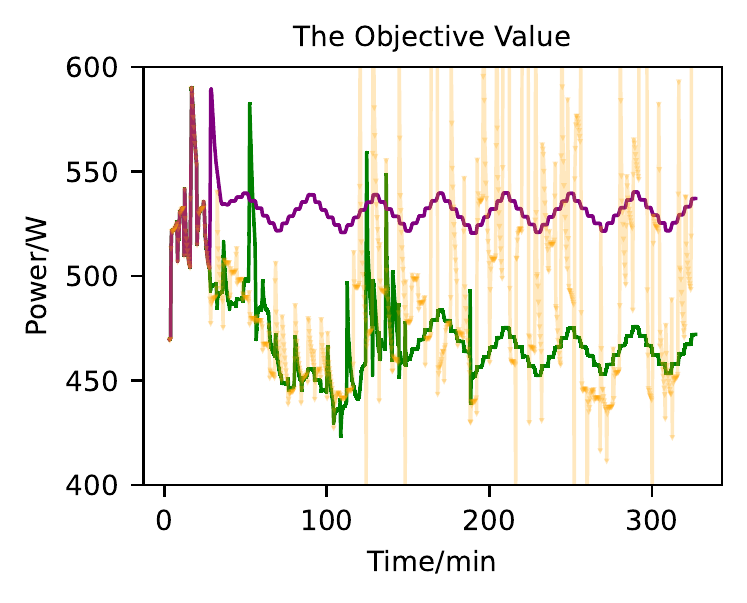}
    \label{fig:recurr_power}
\end{subfigure}
\begin{subfigure}[h]{0.5\columnwidth}
    \centering
    \includegraphics{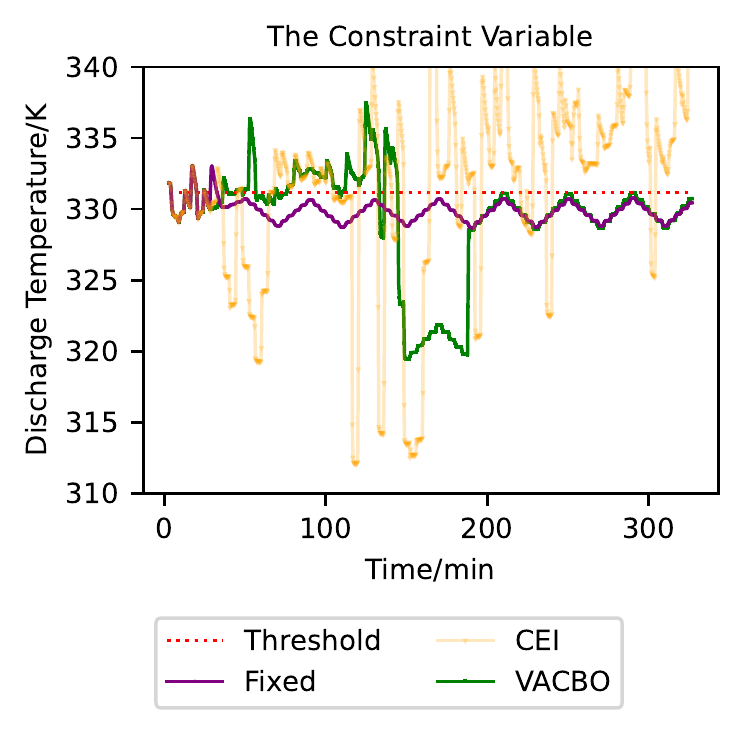}
    \label{fig:recurr_discharge_tmp}
\end{subfigure}
\caption{The evolution of power and discharge temperature with respect to time under recurring contexts.}
\label{fig:recurr_obj_constr}
\end{figure}
\label{sec:art_rec_cts}
\revJnl{To showcase the performance improvement brought by our method, we first apply an artificial sequence of recurring contexts. We periodically choose context $z_t$ from a pre-defined context list $(z^p_i)_{i\in [n]}$ with additive Gaussian noise. Fig.~\ref{fig:recur_context} shows the change of the two contextual variables with respect to time in our experiment.

} 

\subsubsection*{Results and Discussion}
%

Fig.~\ref{fig:recurr_obj_constr} illustrates the experimental results of violation-aware contextual Bayesian optimization compared to a fixed solution\revJnl{, which fixes the tuning parameters to default feasible values}.  The fixed solution can maintain the feasibility of discharge temperature constraint all the time. However, without any set-point exploration and optimization, the fixed solution maintains a relatively high operating power all the time. \revJnl{In} contrast, our VACBO algorithm strategically explores the parameter space and optimizes the power function, with only small short-term tolerable violation. With more and more samples collected, the VACBO algorithm achieves lower and lower power, significantly reducing the energy consumption as compared to the fixed-parameter solution.  
\begin{figure}
    \centering
    \includegraphics{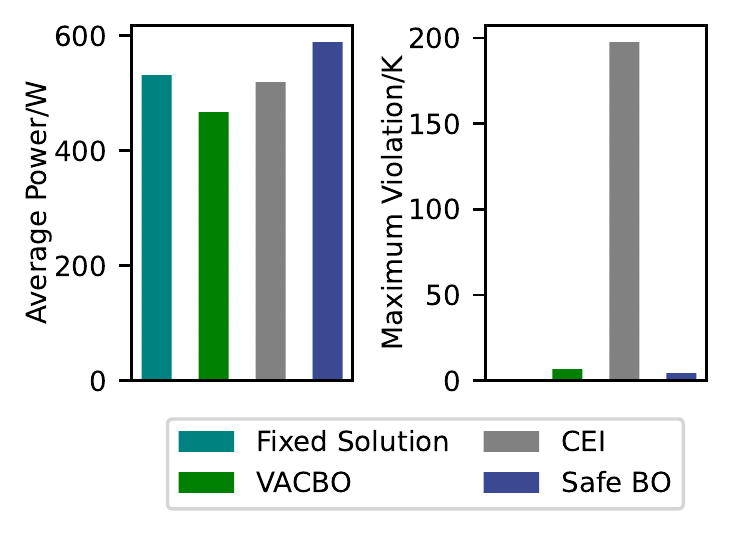}
    \caption{Comparison of average power and maximum constraint violation.}
    \label{fig:recurr_stats}
\end{figure}
\begin{figure}[htbp]
    \centering
    \includegraphics{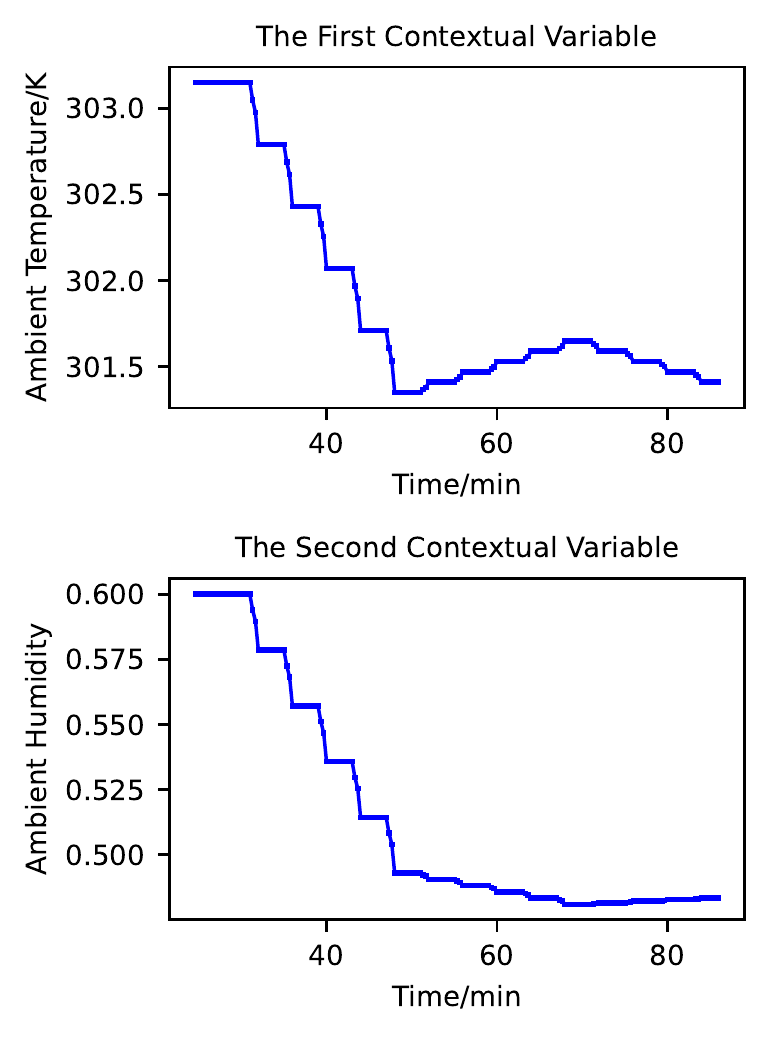}
    \caption{The ambient temperature and the ambient humidity in Zurich, Switzerland, starting at 1:30 $\text{ PM}$ on 2nd, July, 2019.}
    \label{fig:history_context}
\end{figure}

We then compare our VACBO algorithm to two state-of-the-art Bayesian optimization methods with constraints, safe Bayesian optimization~\cite{sui2015safe} and generic constrained Bayesian optimization~\cite{gelbart2014bayesian, gardner2014bayesian}. Fig.~\ref{fig:recurr_stats} gives the comparison of average power and the maximum discharge temperature violation. Our VACBO method reduces the average power by about $12.2\%$ as compared to fixing the setpoints, while managing the discharge temperature well as shown in Fig.~\ref{fig:recurr_obj_constr}. In comparison, generic constrained BO incurs discharge temperature that violates the constraint by as large as more than $150$K, which is very dangerous for the system. As compared to the fixed solution, safe BO shows a very minor improvement in terms of the average power.   


\subsection{Real-world Contexts}

\label{sec:real_cts}
\revJnl{To further demonstrate the effectiveness of our method under real-world contexts, we use the ambient temperature and the ambient humidity in Zurich, Switzerland as the contexts.

Fig.~\ref{fig:history_context} shows the context data used for the real-world contexts. Fig.~\ref{fig:history_obj_constr} shows the evolution of power and discharge temperature with respect to time under real-world contexts for \revJnl{a set of fixed set-points, the generic constrained Bayesian optimization method~(CEI~\cite{gelbart2014bayesian,gardner2014bayesian})}, and VACBO. Similar to the results under recurring contexts, without exploration and optimization of the set-point parameter space, the fixed solution keeps operating with high power. 
We further show the average power and maximum discharge temperature violation in Fig.~\ref{fig:history_stats}. Again, VACBO achieves significant power reduction compared to fixed set-point solution and other state-of-the-art method. Furthermore, our method incurs small and tolerable constraint violation~($<0.5 \textrm{K}$), in sharp contrast to the significant violation~($\geq2.0 \textrm{K}$) of generic contextual constrained BO (CEI method).

\begin{figure}[htbp]
\begin{subfigure}{\columnwidth}
    \centering
    \includegraphics{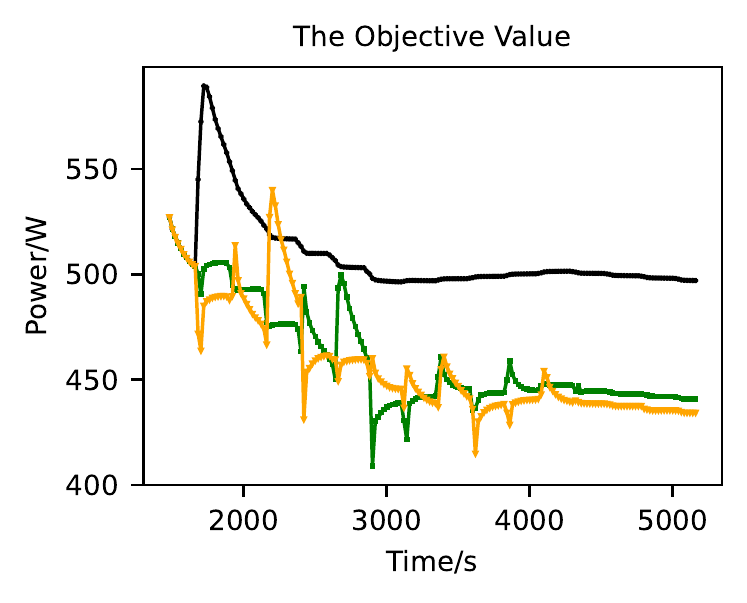}
    \label{fig:history_power}
\end{subfigure}
\begin{subfigure}{\columnwidth}
    \centering
    \includegraphics{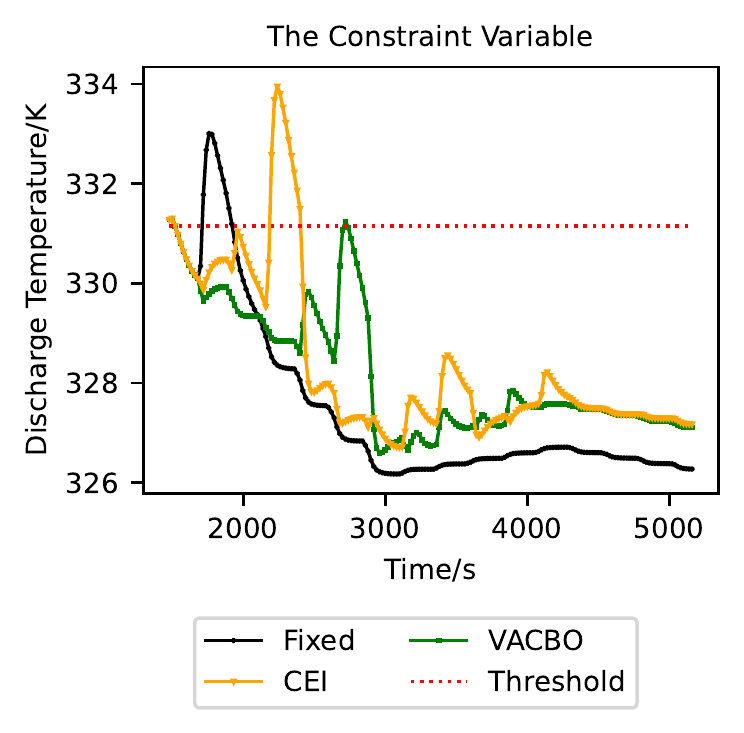}
    \label{fig:history_discharge_tmp}
\end{subfigure}
\caption{The evolution of power and discharge temperature with respect to time under real-world contexts.}
\label{fig:history_obj_constr}
\end{figure}

\begin{figure}[htbp]
    \centering
    \includegraphics{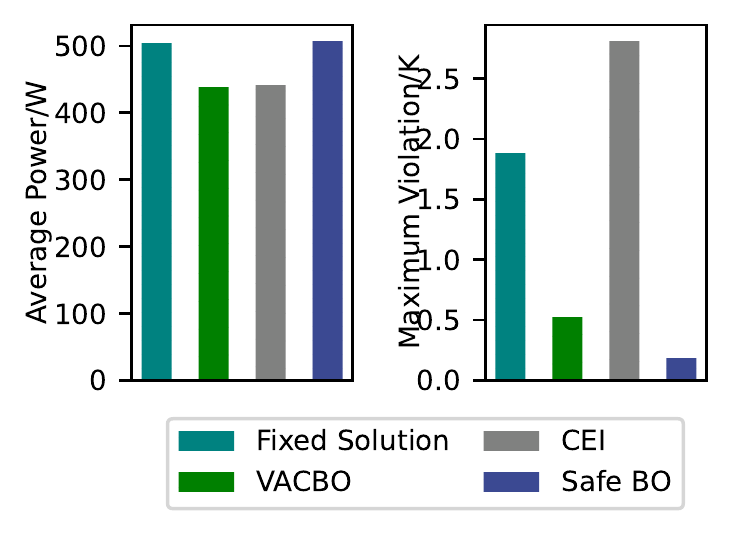}
    \caption{The average power and maximum discharge temperature violation for different methods.}
    \label{fig:history_stats}
\end{figure}

}

\section{Conclusions}
\label{sec:conclusion}
\revJnl{
In this paper, we design a sample-efficient and violation-aware contextual Bayesian optimization~(VACBO) algorithm to solve the closed-loop control performance optimization problem with unmodeled constraints and time-varying contextual factors, by leveraging the fact that small violations over a short period only incur limited costs during the optimization process in many applications such as for vapor-compression cycles. We strategically trade the budgeted violations for faster convergence by solving a tractable auxiliary problem with probabilistic budget constraints at each step. Our experiments on a VCS show that, as compared to existing safe BO and generic constrained BO, our method simultaneously exhibits \revJnl{improved optimization performance} and manages the violation cost well.
}
%

\section*{Acknowledgements}
    
This research was supported by the Swiss National Science Foundation under NCCR Automation, grant agreement {51NF40\_180545}, and in part by the Swiss Data Science Center, grant agreement C20-13.

\bibliographystyle{plain}

\bibliography{ref}

\end{document}